\pdfoutput=1

\documentclass[11pt]{article}

\usepackage{naacl2021}

\usepackage{times}
\usepackage{latexsym}

\usepackage[T1]{fontenc}

\usepackage[utf8]{inputenc}

\usepackage{microtype}

\usepackage{amsmath}
\usepackage{amssymb}
\usepackage{color}
\usepackage{graphicx}
\usepackage{hhline}
\usepackage{makecell}
\usepackage{multirow}
\usepackage{soul}
\usepackage{xcolor}

\DeclareMathOperator*{\argmax}{arg\,max}

\title{Improving Factual Completeness and Consistency of \\ Image-to-Text Radiology Report Generation}

\author{Yasuhide Miura, Yuhao Zhang, Emily Bao Tsai, Curtis P. Langlotz, Dan Jurafsky \\
  Stanford University \\
  \texttt{\{ysmiura, zyh, ebtsai, langlotz, jurafsky\}@stanford.edu}
  }

\begin{document}
\maketitle
\begin{abstract}
Neural image-to-text radiology report generation systems offer the potential to improve radiology reporting by reducing the repetitive process of report drafting and identifying possible medical errors.
However, existing report generation systems, despite achieving high performances on natural language generation metrics such as CIDEr or BLEU, still suffer from incomplete and inconsistent generations. 
Here we introduce two new simple rewards to encourage the generation of factually complete and consistent radiology reports: one that
encourages the system to generate radiology domain \textbf{entities} consistent with the reference,
and one that uses natural language inference to encourage these entities to be described in \textbf{inferentially consistent} ways.
We combine these with the novel use of an existing semantic equivalence metric (BERTScore).
We further propose a report generation system that optimizes these rewards via reinforcement learning. 
On two open radiology report datasets, our system substantially improved the ${\rm F}_1$ score of a clinical information extraction performance by $+22.1$ ($\Delta +63.9\%$). 
We further show via a human evaluation and a qualitative analysis that our system leads to generations that are more factually complete and consistent compared to the baselines.
\end{abstract}

\section{Introduction}

\begin{figure}[t]
\centering
\includegraphics[width=1.0\columnwidth]{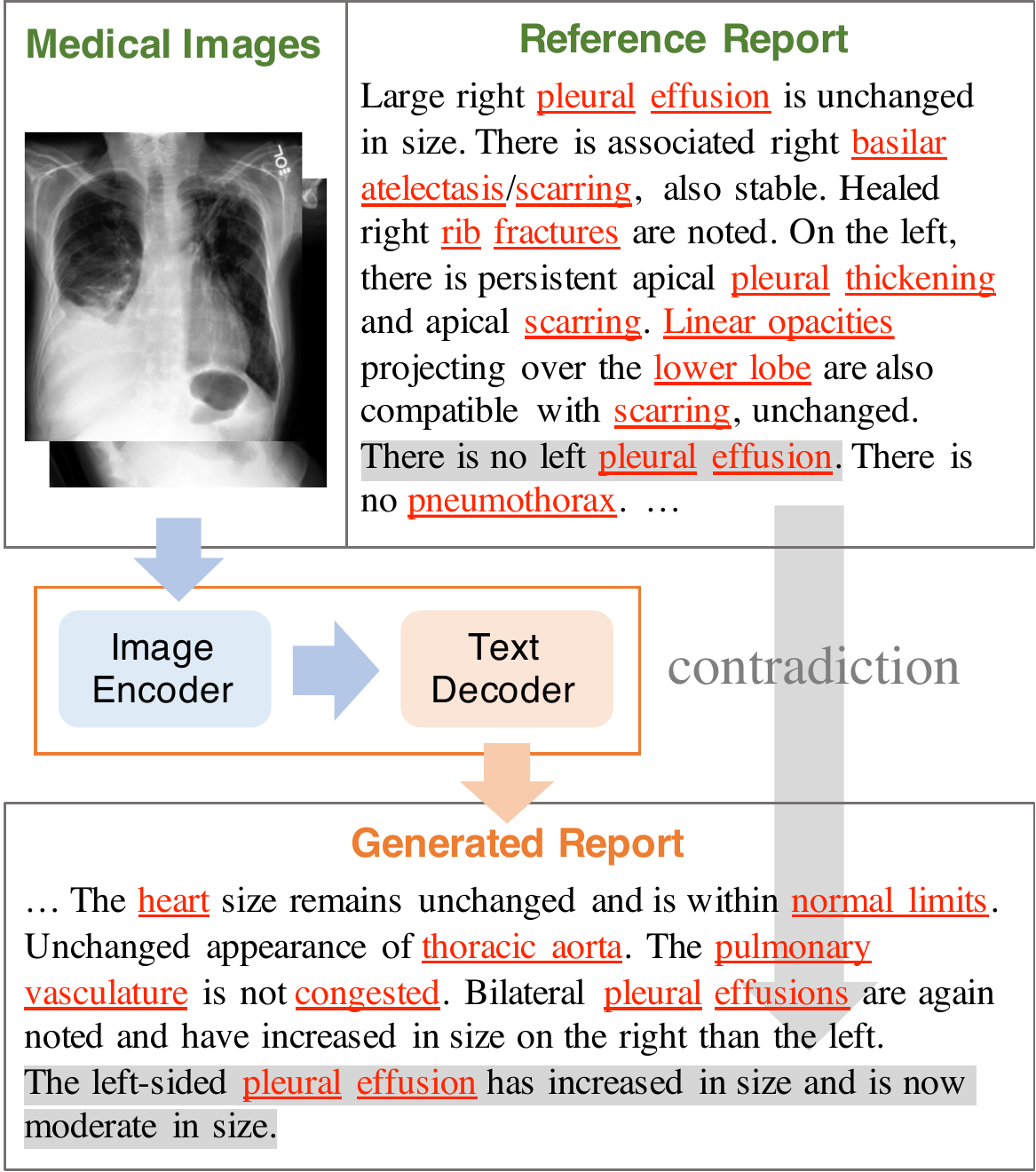} 
\caption{A (partial) example of a report generated from our system (with ``\ldots'' representing abbreviated text). The system encodes images and generates text from that encoded representation. Underlined words are disease and anatomy entities. The shaded sentences are an example of a contradictory pair.}
\label{fig:radrep}
\end{figure}

An important new application of natural language generation (NLG) is to
build assistive systems that take X-ray images of a patient
and generate a textual report describing clinical observations in the images
\cite{Jing2018,Li2018,Liu2019,Boag2020,Chen2020}.
Figure \ref{fig:radrep} shows an example of a radiology report generated by such a system.
This is a clinically important task,  offering the potential to reduce
radiologists' repetitive work and generally improve clinical communication \cite{Kahn2009}.

Automatic radiology report generation systems have achieved promising performance as measured by widely used NLG metrics such as CIDEr \cite{Vedantam2015cider} and BLEU \cite{Papineni2002} on several datasets \cite{Li2018,Jing2019,Chen2020}.
However, reports that achieve high performance on these NLG metrics are not always factually complete or consistent.
In addition to the use of inadequate metrics, the factual incompleteness and inconsistency issue in generated reports is further exacerbated by the inadequate training of these systems.
Specifically, the standard teacher-forcing training algorithm \cite{Williams1989} used by most existing work can lead to a discrepancy between what the model sees during training and test time \cite{Ranzato2016}, resulting in degenerate outputs with factual hallucinations \cite{Maynez2020}.
\citet{Liu2019} and \citet{Boag2020} have shown that reports generated by state-of-the-art systems still have poor quality when evaluated by their clinical metrics as measured with an information extraction system designed for radiology reports.
For example, the generated report in Figure \ref{fig:radrep} is incomplete since it neglects an observation of \textit{atelectasis} that can be found in the images.
It is also inconsistent since it mentions \textit{left-sided pleural effusion} which is not present in the images.
Indeed, we show that existing systems are inadequate in factual completeness and consistency, and that an image-to-text radiology report generation system can be substantially improved by replacing widely used NLG metrics with simple alternatives.

We propose two new simple rewards that can encourage the factual completeness and consistency of the generated reports.
First, we propose the Exact Entity Match Reward ($\mathrm{fact}_\mathrm{ENT}$) which captures the completeness of a generated report by measuring its coverage of entities in the radiology domain, compared with a reference report.
The goal of the reward is to better capture disease and anatomical knowledge that are encoded in the entities.
Second, we propose the Entailing Entity Match Reward ($\mathrm{fact}_\mathrm{ENTNLI}$), which extends ${\rm fact_{ENT}}$ with a natural language inference (NLI) model that further considers how inferentially consistent the generated entities are with their descriptions in the reference.
We add NLI to control the overestimation of disease when optimizing towards $\mathrm{fact}_\mathrm{ENT}$.
We use these two metrics along with an existing semantic equivalence metric, BERTScore \cite{Zhang2020bertscore}, to potentially capture synonyms (e.g., “left and right” effusions are synonymous with “bilateral” effusions) and distant dependencies between diseases (e.g., a negation like “\ldots but underlying consolidation or other pulmonary lesion not excluded”) that are present in radiology reports.

Although recent work in summarization, dialogue, and data-to-text generation has tried to address this problem of factual incompleteness and inconsistency  by using natural language inference (NLI) \cite{Falke2019, Welleck2019}, question answering (QA) \cite{Wang2020a}, or content matching constraint \cite{Wang2020b} approaches, they either show negative results or are not directly applicable to the generation of radiology reports due to a substantial task and domain difference.
To construct the NLI model for $\mathrm{fact}_\mathrm{ENTNLI}$, we present a weakly supervised approach that adapts an existing NLI model to the radiology domain.
We further present a report generation model which directly optimizes a Transformer-based architecture with these rewards using reinforcement learning (RL).

We evaluate our proposed report generation model on two publicly available radiology report generation datasets. 
We find that optimizing the proposed rewards along with BERTScore by RL leads to generated reports that achieve substantially improved performance in the important clinical metrics \cite{Liu2019,Boag2020,Chen2020}, demonstrating the higher clinical value of our approach.
We make all our code and the expert-labeled test set for evaluating the radiology NLI model publicly available to encourage future research\footnote{\url{https://github.com/ysmiura/ifcc}}.
To summarize, our contributions in this paper are:
\vspace*{-5pt}
\begin{enumerate}
\setlength{\itemsep}{0pt}
\setlength{\parskip}{0pt}
\item We propose two simple rewards for image-to-text radiology report generation, which focus on capturing the factual completeness and consistency of generated reports, and a weak supervision-based approach for training a radiology-domain NLI model to realize the second reward.
\item We present a new radiology report generation model that directly optimizes these new rewards with RL, 
showing that previous approaches that optimize traditional NLG metrics are inadequate,
and that the proposed approach substantially improves performance on clinical metrics (as much as $\Delta +64.2\%$) on two publicly available datasets.
\end{enumerate}

\section{Related Work}
\subsection{Image-to-Text Radiology Report Generation}

\citet{Wang2018} and \citet{Jing2018} first proposed multi-task learning models that jointly generate a report and classify disease labels from a chest X-ray image.
Their models were extended to use multiple images \cite{Yuan2019}, to adopt a hybrid retrieval-generation model \cite{Li2018}, or to consider structure information \cite{Jing2019}.
More recent work has focused on generating reports that are clinically consistent and accurate.
\citet{Liu2019} presented a system that generates accurate reports by fine-tuning it with their Clinically Coherent Reward.
\citet{Boag2020} evaluated several baseline generation systems with clinical metrics and found that standard NLG metrics are ill-equipped for this task.
Very recently, \citet{Chen2020} proposed an approach to generate radiology reports with a memory-driven Transformer.
Our work is most related to \citet{Liu2019}; their system, however, is dependent on a rule-based information extraction system specifically created for chest X-ray reports and has limited robustness and generalizability to different domains within radiology.
By contrast, we aim to develop methods that improve the factual completeness and consistency of generated reports by harnessing more robust statistical models and are easily generalizable.

\subsection{Consistency and Faithfulness in Natural Language Generation}

A variety of recent work has focused on consistency and faithfulness in generation.
Our work is inspired by  \citet{Falke2019}, \citet{Welleck2019}, and \citet{Matsumaru2020} in using NLI to rerank or filter generations in text summarization, dialogue, and headline generations systems, respectively.
Other attempts in this direction include evaluating consistency in generations using QA models \cite{Durmus2020, Wang2020a, Maynez2020}, with distantly supervised classifiers \cite{Kryscinski2020}, and with task-specific content matching constraints \cite{Wang2020b}.
\citet{Liu2019} and \citet{Zhang2020a} studied improving the factual correctness in generating radiology reports with rule-based information extraction systems.
Our work mainly differs from theirs in the direct optimization of factual completeness with an entity-based reward and of factual consistency with a statistical NLI-based reward.

\subsection{Image Captioning with Transformer}
The problem of generating text from image data has been widely studied in the image captioning setting.
While early work  focused on combining convolutional neural network (CNN) and recurrent neural network (RNN) architectures \cite{Vinyals2015}, more recent work has discovered the effectiveness of using the Transformer architecture \cite{Vaswani2017}.
\citet{Li2019} and \citet{Pan2020} introduced an attention process to exploit semantic and visual information into this architecture.
\citet{Herdade2019}, \citet{Cornia2020}, and \citet{Guo2020} extended this architecture to learn geometrical and other relationships between input regions.
We find Meshed-Memory Transformer \cite{Cornia2020} ($\mathcal{M}^2$ Trans) to be more effective in our radiology report generation task than the traditional RNN-based models and Transformer models (an empirical result will be shown in \S\ref{sec:exp}), and therefore use it as our base architecture.

\section{Methods}

\subsection{Image-to-Text Radiology Report Generation with $\boldsymbol{\mathcal{M}^2}$ Trans}
Formally, given $K$ individual images $x_{1 \ldots K}$ of a patient, our task involves generating a sequence of words to form a textual report $\hat{y}$, which describes the clinical observations in the images.
This task resembles  image captioning, except with multiple images as  input and longer text sequences as output.
We therefore extend a state-of-the-art image captioning model, $\mathcal{M}^2$ Trans \cite{Cornia2020}, with multi-image input as our base architecture.
We first briefly introduce this model and refer interested readers to \citet{Cornia2020}.

\begin{figure}[t]
\centering
\includegraphics[width=0.9\columnwidth]{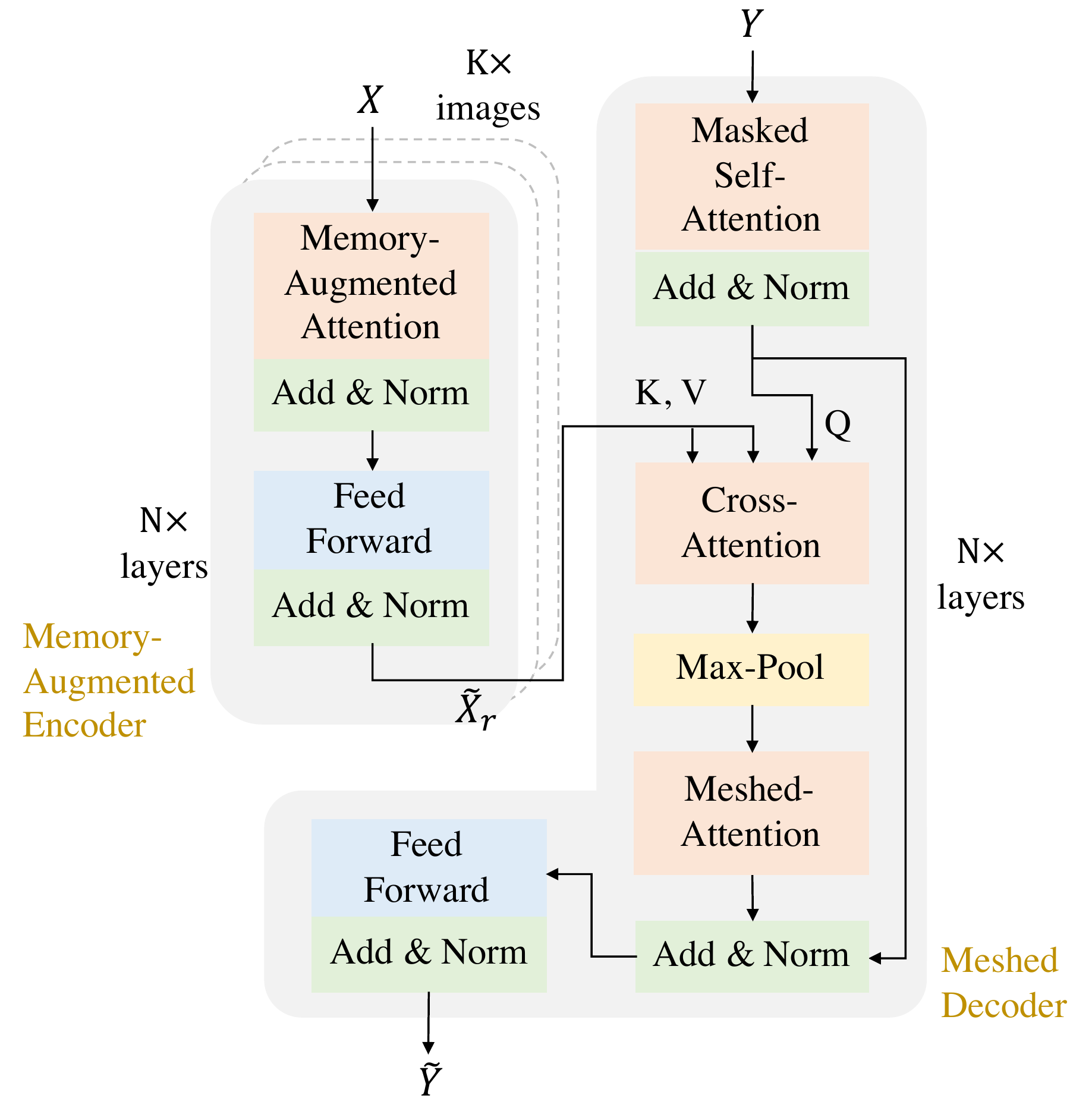} 
\caption{An overview of Meshed-Memory Transformer extended to multiple images.}
\label{fig:m2trans}
\end{figure}

Figure \ref{fig:m2trans} illustrates an overview of the $\mathcal{M}^2$ Trans model.
Given an image $x_k$, image regions are first extracted with a CNN as $\boldsymbol{X} = \mathrm{CNN}(x_k)$.
$\boldsymbol{X}$ is then encoded with a memory-augmented attention process $\mathcal{M}_\mathrm{mem}(\boldsymbol{X})$ as

\vspace*{-10pt}
{\small
\begin{align}
\mathcal{M}_{\rm{mem}}(\boldsymbol{X}) & = \rm{Att}(W_q \boldsymbol{X}, \boldsymbol{K}, \boldsymbol{V}) \\
\rm{Att}(\boldsymbol{Q}, \boldsymbol{K}, \boldsymbol{V}) & = \mathrm{softmax} \left( \frac{\boldsymbol{Q} \boldsymbol{K}^T}{\sqrt{d}} \right) \boldsymbol{V} \\
\boldsymbol{K} & = [ W_k \boldsymbol{X}; \boldsymbol{M}_k ] \\
\boldsymbol{V} & = [ W_v \boldsymbol{X}; \boldsymbol{M}_v]
\end{align}
}%
where $W_q, W_k, W_v$ are weights, $\boldsymbol{M}_k, \boldsymbol{M}_v$ are memory matrices, $d$ is a scaling factor, and $[* ; *]$ is the concatenation operation.
$\rm{Att}(\boldsymbol{Q}, \boldsymbol{K}, \boldsymbol{V})$ is an attention process derived from the Transformer architecture \cite{Vaswani2017} and extended to include  memory matrices that can encode a priori knowledge between image regions.
In the encoder, this attention process is a self-attention process since all of the query $\boldsymbol{Q}$, the key $\boldsymbol{K}$, and the value $\boldsymbol{V}$ depend on $\boldsymbol{X}$. 
$\mathcal{M}_\mathrm{mem}(\boldsymbol{X})$ is further processed with a feed forward layer, a residual connection, and a layer normalization to output $\tilde{\boldsymbol{X}}$.
This encoding process can be stacked $N$ times and is applied to $K$ images, and $n$-th layer output of $K$ image will be $\tilde{\boldsymbol{X}}_{n, \boldsymbol{K}}$.

The meshed decoder first processes an encoded text $\boldsymbol{Y}$ with a masked self-attention and further processes it with a feed forward layer, a residual connection, and a layer normalization to output $\ddot{\boldsymbol{Y}}$.
$\ddot{\boldsymbol{Y}}$ is then passed to a cross attention $\mathcal{C}(\tilde{\boldsymbol{X}}_{n, \boldsymbol{K}}, \ddot{\boldsymbol{Y}})$ and a meshed attention $\mathcal{M}_\mathrm{mesh}(\tilde{\boldsymbol{X}}_{\boldsymbol{N}, \boldsymbol{K}}, \ddot{\boldsymbol{Y}})$ as 

\vspace*{-10pt}
{\small
\begin{gather}
\mathcal{M}_\mathrm{mesh} (\tilde{\boldsymbol{X}}_{\boldsymbol{N}, \boldsymbol{K}}, \ddot{\boldsymbol{Y}}) = \sum_n \boldsymbol{\alpha}_n \odot \mathcal{C}(\tilde{\boldsymbol{X}}_{n, \boldsymbol{K}}, \ddot{\boldsymbol{Y}}) \\
\mathcal{C}(\tilde{\boldsymbol{X}}_{n, \boldsymbol{K}}, \ddot{\boldsymbol{Y}}) = \max_{\boldsymbol{K}} (\rm{Att}(W_q \ddot{\boldsymbol{Y}}, W_k \tilde{\boldsymbol{X}}_{n, \boldsymbol{K}}, W_v \tilde{\boldsymbol{X}}_{n, \boldsymbol{K}})) \\
\boldsymbol{\alpha}_n = \sigma \left( W_n [ \boldsymbol{Y} ;  \mathcal{C}(\tilde{\boldsymbol{X}}_{n, \boldsymbol{K}}, \ddot{\boldsymbol{Y}})] + b_n \right)
\end{gather}
}%
where $\odot$ is element-wise multiplication, $\max_{\boldsymbol{K}}$ is max-pooling over $K$ images, $\sigma$ is sigmoid function, $W_n$ is a weight, and $b_n$ is a bias.
The weighted summation in $\mathcal{M}_{\rm mesh}(\tilde{\boldsymbol{X}}_{\boldsymbol{N}, \boldsymbol{K}}, \ddot{\boldsymbol{Y}})$ exploits both low-level and high-level information from the $N$ stacked encoder.
Differing from the self-attention process in the encoder, the cross attention uses a query that depends on $\boldsymbol{Y}$ and a key and a value that depend on $\boldsymbol{X}$.
$\mathcal{M}_\mathrm{mesh} (\tilde{\boldsymbol{X}}_{\boldsymbol{N}, \boldsymbol{K}}, \ddot{\boldsymbol{Y}})$ is further processed with a feed forward layer, a residual connection, and a layer normalization to output $\tilde{\boldsymbol{Y}}$.
As like in the encoder, the decoder can be stacked $N$ times to output $\tilde{\boldsymbol{Y}}_N$.
$\tilde{\boldsymbol{Y}}_N$ is further passed to a feed forward layer to output report $\hat{y}$.

\subsection{Optimization with Factual Completeness and Consistency}

\subsubsection{Exact Entity Match Reward ($\boldsymbol{\mathrm{fact}_\mathrm{ENT}}$)}

We designed an F-score entity match reward to capture factual completeness.
This reward assumes that entities encode disease and anatomical knowledge that relates to factual completeness.
A named entity recognizer is applied to $\hat{y}$ and the corresponding reference report $y$.
Given entities $E_\mathrm{gen}$ and $E_\mathrm{ref}$ recognized from $y_\mathrm{gen}$ and $y_\mathrm{ref}$ respectively, precision (pr) and recall (rc) of entity match are calculated as 

\vspace*{-10pt}
{\small
\begin{align}
\mathrm{pr}_\mathrm{ENT} & = \frac{\sum_{e \in E_\mathrm{gen}} \delta (e, E_\mathrm{ref}) }{| E_\mathrm{gen} |} \\
\mathrm{rc}_\mathrm{ENT} & = \frac{\sum_{e\in E_\mathrm{ref}} \delta (e, E_\mathrm{gen}) }{| E_\mathrm{ref} |} \\
\delta (e, E) & = 
\begin{cases}
1 ,& \text{\small for } e \in E \\
0 ,& \text{\small otherwise} 
\end{cases} \label{eq:eem-delta}
\end{align}
}%
The harmonic mean of precision and recall is taken as $\mathrm{fact}_\mathrm{ENT}$ to reward a balanced match of entities.
We used Stanza \cite{Qi2020} and its clinical models \cite{Zhang2020c} as a named entity recognizer for radiology reports.
For example in the case of Figure \ref{fig:radrep}, the common entities among the reference report and the generated report are \textit{pleural} and \textit{effusion}, resulting to $\mathrm{fact}_\mathrm{ENT} = 33.3$.

\subsubsection{Entailing Entity Match Reward ($\boldsymbol{\mathrm{fact}_\mathrm{ENTNLI}}$)}

We additionally designed an F-score style reward that expands $\mathrm{fact}_\mathrm{ENT}$ with NLI to capture factual consistency.
NLI is used to control the overestimation of disease when optimizing towards $\mathrm{fact}_\mathrm{ENT}$.
In $\mathrm{fact}_\mathrm{ENTNLI}$, $\delta$ in Eq. \ref{eq:eem-delta} is expanded to 

\vspace*{-10pt}
{\small
\begin{align}
\scriptstyle \phi (e, E) = 
\begin{cases}
1 ,& \text{\small for } e \in E  \wedge \mathrm{NLI_e}(\boldsymbol{P}, h) \ne \mathrm{contradiction} \\
1 ,& \text{\small for } \mathrm{NLI_e}(\boldsymbol{P}, h) = \mathrm{entailment} \\ 
0 ,& \text{\small otherwise} 
\end{cases} \\
{\rm NLI_e}(\boldsymbol{P}, h) = nli(\hat{p}, h) \text{ where } \hat{p} = \argmax_{p \in \boldsymbol{P}} sim(h, p)
\end{align}
}%
where $h$ is a sentence that includes $e$, $\boldsymbol{P}$ is all sentences in a counter part text (if $h$ is a sentence in a generated report, $\boldsymbol{P}$ is all sentences in the corresponding reference report), $nli(*, *)$ is an NLI function that returns an NLI label which is one of $\{\mathrm{entailment, neutral, contradiction}\}$, and $sim (*, *)$ is a text similarity function.
We used BERTScore \cite{Zhang2020bertscore} as $sim (*, *)$ in the experiments (the detail of BERTScore can be found in Appendix \ref{sec:app-nli}).
The harmonic mean of precision and recall is taken as $\mathrm{fact}_\mathrm{ENTNLI}$ to encourage a balanced factual consistency between a generated text and the corresponding reference text.
For example in the case of Figure \ref{fig:radrep}, the sentence ``The left-sided pleural effusion has increased in size and is now moderate in size.'' will be contradictory to ``There is no left pleural effusion.'' resulting in \textit{pleural} and \textit{effusion} being rejected in $y_\mathrm{gen}$.

\subsubsection{Joint Loss for Optimizing Factual Completeness and Consistency}

We integrate the proposed factual rewards into self-critical sequence training \cite{Rennie2017}.
An RL loss $\mathcal{L_\mathrm{RL}}$ is minimized as the negative expectation of the reward $r$.
The gradient of the loss is estimated with a single Monte Carlo sample as

\vspace*{-10pt}
{\small
\begin{align}
\nabla_{\theta} \mathcal{L}_\mathrm{RL} (\theta) = - \nabla_{\theta} \log P_{\theta} (  \hat{y}_{sp} | x_{1 \ldots K}) \left( r( \hat{y}_{sp} ) - r( \hat{y}_{gd} ) \right)
\end{align}
}%
where $\hat{y}_{sp}$ is a sampled text and $\hat{y}_{gd}$ is a greedy decoded text.
\citeauthor{Paulus2018} \shortcite{Paulus2018} and \citeauthor{Zhang2020a} \shortcite{Zhang2020a} have shown that a generation can be improved by combining multiple losses.
We combine a factual metric loss with a language model loss and an NLG loss as 

\vspace*{-10pt}
{\small
\begin{align}
\mathcal{L} & = \lambda_1 \mathcal{L}_\mathrm{NLL} + \lambda_2 \mathcal{L}_\mathrm{RL\_NLG} + \lambda_3 \mathcal{L}_\mathrm{RL\_FACT} 
\end{align}
}%
where $\mathcal{L}_\mathrm{NLL}$ is a language model loss, $\mathcal{L}_\mathrm{RL\_NLG}$ is the RL loss using an NLG metric (e.g., CIDEr or BERTScore), $\mathcal{L}_\mathrm{RL\_FACT}$ is the RL loss using a  factual reward (e.g., $\mathrm{fact}_\mathrm{ENT}$ or $\mathrm{fact}_\mathrm{ENTNLI}$), and $\lambda_{*}$ are scaling factors to balance the multiple losses.

\subsection{A Weakly-Supervised Approach for Radiology NLI}

We propose a weakly-supervised approach to construct an NLI model for radiology reports.
(There already exists an NLI system for the medical domain, MedNLI \cite{Romanov2018}, but we found that a model trained on MedNLI does not work well on radiology reports.)
Given a large scale dataset of radiology reports, a sentence pair is sampled and filtered with weakly-supervised rules.
The rules are prepared to extract a randomly sampled sentence pair ($s_1$ and $s_2$) that are in an entailment, neutral, or contradiction relation.
We designed the following $6$ rules for weak-supervision.
\begin{description}
\item[Entailment 1 (E1)]
(1) $s_1$ and $s_2$ are semantically similar and
(2) NE of $s_2$ is a subset or equal to NE of $s_1$.
\item[Neutral 1 (N1)]
(1) $s_1$ and $s_2$ are semantically similar and
(2) NE of $s_1$ is a subset of NE of $s_2$.
\item[Neutral 2 (N2)]
(1) NE of $s_1$ are equal to NE of $s_2$ and
(2) $s_1$ include an antonym of a word in $s_2$.
\item[Neutral 3 (N3)]
(1) NE types of $s_1$ are equal to NE types of $s_2$ and
(2) NE of $s_1$  is different from NE of $s_2$.
NE types are used in this rule to introduce a certain level of similarity between $s_1$ and $s_2$.
\item[Neutral 4 (N4)]
(1) NE of $s_1$ are equal to NE of $s_2$ and
(2) $s_1$ and $s_2$ include observation keywords.
\item[Contradiction 1 (C1)]
(1) NE of $s_1$ is equal or a subset to NE of $s_2$ and
(2) $s_1$ is a negation of $s_2$.
\end{description}
The rules rely on a semantic similarity measure and the overlap of entities to determine the relationship between $s_1$ and $s_2$.
In the neutral rules and the contradiction rule, we included similarity measures to avoid extracting easy to distinguish sentence pairs.

We evaluated this NLI by preparing training data, validation data, and test data.
For the training data, the training set of MIMIC-CXR \cite{Johnson2019} is used as the source of sentence pairs.
$2$k pairs are extracted for E1 and C1, $0.5$k pairs are extracted for N1, N2, N3, and N4, resulting in a total of $6$k pairs.
The training set of MedNLI is also used as additional data.
For the validation data and the test data, we sampled $480$ sentence pairs from the validation section of MIMIC-CXR and had them annotated by two experts: one medical expert and one NLP expert.
Each pair is annotated twice swapping its premise and hypothesis resulting in $960$ pairs and are split in half resulting in $480$ pair for a validation set and $480$ pairs for a test set.
The test set of MedNLI is also used as alternative test data.

\begin{table}[t]
\small
\centering
\def\arraystretch{1.1}
\begin{tabular}{ lc | c c}
\hline
\multirow{2}{*}{Training Data} & \multirow{2}{*}{\#samples} & \multicolumn{2}{c}{Test Accuracy} \\
& & RadNLI & MedNLI \\
\hhline{==|==}
MedNLI & $13$k & $53.3$ & $\boldsymbol{80.9}$ \\
MedNLI + RadNLI & $19$k & $\boldsymbol{77.8}$ &  $79.8$ \\
\hline
\end{tabular}
\caption{The accuracies of the NLI model trained with the weakly-supervised approach. RadNLI is the proposed NLI for radiology reports. The values are the average of $5$ runs and the bold values are the best results of each test set.}
\label{tab:radnli}
\end{table}

We used BERT \cite{Devlin2019} as an NLI model since it performed as a strong baseline in the existing MedNLI system \cite{BenAbacha2019}, and used Stanza \cite{Qi2020} and its clinical models \cite{Zhang2020c} as a named entity recognizer.
Table \ref{tab:radnli} shows the result of the model trained with and without the weakly-supervised data.
The accuracy of NLI on radiology data increased substantially by $+24.5$\% with the addition of the radiology NLI training set.
(See Appendix \ref{sec:app-nli} for the detail of the rules, the datasets, and the model configuration.)

\section{Experiments}
\label{sec:exp}

\subsection{Data}

We used the training and validation sets of MIMIC-CXR \cite{Johnson2019} to train and validate models.
MIMIC-CXR is a large publicly available database of chest radiographs.
We extracted the \textit{findings} sections from the reports with a text extraction tool for MIMIC-CXR\footnote{https://github.com/MIT-LCP/mimic-cxr/tree/master/txt}, and used them as our reference reports as in previous work \cite{Liu2019, Boag2020}.
\textit{Findings} section is a natural language description of the important aspects in a radiology image.
The reports with empty \textit{findings} sections were discarded, resulting in $152173$ and $1196$ reports for the training and validation set, respectively.
We used the test set of MIMIC-CXR and the entire Open-i Chest X-ray dataset \cite{Demner-Fushman2012} as two individual test sets.
Open-i is another publicly available database of chest radiographs which has been widely used in past studies.
We again extracted the \textit{findings} sections, resulting in $ 2347$ reports for MIMIC-CXR and $3335$ reports for Open-i.
Open-i is used only for testing since the number of reports is too small to train and test a neural report generation model.

\subsection{Evaluation Metrics}
\label{sec:metrics}

\textbf{BLEU4, CIDEr-D \& BERTScore: }
We first use general NLG metrics to evaluate the generation quality. 
These metrics include the 4-gram BLEU scroe \cite[BLEU4]{Papineni2002}, CIDEr score \cite{Vedantam2015cider} with gaming penalties (CIDEr-D), and the $\rm{F}_1$ score of the BERTScore \cite{Zhang2020bertscore}. \\
\textbf{Clinical Metrics: }
However, NLG metrics such as BLEU and CIDEr are known to be inadequate for evaluating factual completeness and consistency.
We therefore followed previous work \cite{Liu2019, Boag2020, Chen2020} by  additionally evaluating the clinical accuracy of the generated reports using a clinical information extraction system.
We use CheXbert \cite{Smit2020}, an information extraction system for chest reports, to extract the presence status of a series of observations (i.e., whether a disease is present or not), and score a generation by comparing the values of these observations to those obtained from the reference\footnote{We used CheXbert instead of CheXpert \cite{Irvin2019} since CheXbert was evaluated to be approximately $5.5\%$ more accurate than CheXpert. The evaluation using CheXpert can be found in Appendix \ref{sec:app-cfm}.}.
The micro average of accuracy, precision, recall, and $\rm{F}_1$ scores are calculated over $5$ observations (following previous work \cite{Irvin2019}) for: \textit{atelectasis}, \textit{cardiomegaly}, \textit{consolidation}, \textit{edema}, and \textit{pleural effusion}\footnote{These $5$ observations are evaluated to be most represented in real-world radiology reports and therefore using these 5 observations (and excluding others) leads to less variance and more statistical strength in the results. We include the detailed results of the clinical metrics in Appendix C for completeness.
}. \\
\textbf{$\boldsymbol{\mathrm{fact}_\mathrm{ENT}}$ \& $\boldsymbol{\mathrm{fact}_\mathrm{ENTNLI}}$:}
We additionally include our proposed rewards $\rm fact_{ENT}$ and $\rm fact_{ENTNLI}$ as metrics to compare their values for different models.

\subsection{Model Variations}

\begin{table*}[t]
\small
\centering
\def\arraystretch{1.1} 
\begin{tabular}{ l | l c c c c c c c c c}
\hline
\multirow{2}{*}{Dataset} & \multirow{2}{*}{Model} & \multicolumn{3}{c}{NLG Metrics} & \multicolumn{4}{c}{Clinical Metrics (micro-avg)} & \multicolumn{2}{c}{Factual Rewards} \\
& & BL4 & CDr & BS &  P & R & $\mathrm{F}_1$ & acc. & $\mathrm{fc}_\mathrm{E}$ & $\mathrm{fc}_\mathrm{EN}$ \\
\hhline{=|==========}
& \multicolumn{10}{l}{\textit{Previous models}} \\
& TieNet \cite{Wang2018} & $8.1$ & $37.2$ & $49.2$ & $38.6$ & $20.9$ & $27.1$ & $74.0$ & $-$ & $-$ \\
& $\mathrm{CNN}$-$\mathrm{RNN^2}$ \cite{Liu2019} & 7.6 & 44.7 & 41.2 & $\boldsymbol{66.4}$ & $18.7$ & $29.2$ & $\boldsymbol{79.0}$ & $-$ & $-$ \\
& R2Gen \cite{Chen2020} & $8.6$ & $40.6$ & $50.8$  & $41.2$ & $29.8$ & $34.6$ & $73.9$ & $-$ & $-$ \\
\cline{2-11}
\multirow{2}{*}{MIMIC-}& \multicolumn{10}{l}{\textit{Proposed approach without proposed optimization}} \\
\multirow{2}{*}{CXR} & $\mathcal{M}^2$ Trans w/ NLL & $10.5$ & $44.5$ & $51.2$ & $48.9$ & $41.1$ & $44.7$ & $76.5$ & $27.3$ & $24.4$ \\
& $\mathcal{M}^2$ Trans w/ NLL+CDr & $\boldsymbol{13.3}$ & $\boldsymbol{67.0}$ & $55.9$ & $50.0$ & $51.3$ & $50.6$ & $76.9$ & $35.2$ & $32.9$ \\
\cline{2-11}
& \multicolumn{10}{l}{\textit{Proposed approach}} \\
& $\mathcal{M}^2$ Trans w/ NLL+BS & $12.2$ & $58.4$ & $\boldsymbol{58.4}$ & $46.3$ & $67.5$ & $54.9$ & $74.4$ & $35.9$ & $33.0$ \\
& $\mathcal{M}^2$ Trans w/ NLL+BS+$\mathrm{fc}_\mathrm{E}$ & $11.1$ & $49.2$ & $57.2$ & $46.3$ & $\boldsymbol{73.2}$ & $\boldsymbol{56.7}$ & $74.2$ & $\boldsymbol{39.5}$ & $34.8$ \\
& $\mathcal{M}^2$ Trans w/ NLL+BS+$\mathrm{fc}_\mathrm{EN}$ & $11.4$ & $50.9$ & $56.9$ & $50.3$ & $65.1$ & $\boldsymbol{56.7}$ & $77.1$ & $38.5$ & $\boldsymbol{37.9}$ \\
\hhline{=|==========}
& \multicolumn{10}{l}{\textit{Previous models}} \\
& TieNet \cite{Wang2018} & $9.0$ & $65.7$ & $56.1$ & $46.9$ & $15.9$ & $23.7$ & $96.0$ & $-$ & $-$ \\
& $\mathrm{CNN}$-$\mathrm{RNN^2}$ \cite{Liu2019} & 12.1 & 87.2 & 57.1 & $\boldsymbol{55.1}$ & $7.5$ & $13.2$ & $\boldsymbol{96.1}$ & $-$ & $-$ \\
& R2Gen \cite{Chen2020} & $6.7$ & $61.4$ & $53.8$ & $27.0$ & $17.3$ & $21.1$ & $94.9$ & $-$ & $-$ \\
\cline{2-11}
& \multicolumn{10}{l}{\textit{Proposed approach without proposed optimization}} \\
Open-i & $\mathcal{M}^2$ Trans w/ NLL & $8.2$ & $64.4$ & $53.1$ & $44.7$ & $32.7$ & $37.8$ & $95.8$ & $31.1$ & $34.1$ \\
& $\mathcal{M}^2$ Trans w/ NLL+CDr & $\boldsymbol{13.4}$ & $97.2$ & $59.9$ & $48.2$ & $24.2$ & $32.2$ & $96.0$ & $40.6$ & $42.9$ \\
\cline{2-11}
& \multicolumn{10}{l}{\textit{Proposed approach}} \\
& $\mathcal{M}^2$ Trans w/ NLL+BS & $12.3$ & $87.3$ & $62.4$ & $47.7$ & $46.6$ & $47.2$ & $95.9$ & $41.5$ & $44.1$ \\
& $\mathcal{M}^2$ Trans w/ NLL+BS+$\mathrm{fc}_\mathrm{E}$ & $12.0$ & $99.6$ & $\boldsymbol{62.6}$ & $44.0$ & $\boldsymbol{53.5}$ & $\boldsymbol{48.3}$ & $95.5$ & $\boldsymbol{44.4}$ & $46.8$  \\
& $\mathcal{M}^2$ Trans w/ NLL+BS+$\mathrm{fc}_\mathrm{EN}$ & $13.1$ & $\boldsymbol{103.4}$ & $61.0$ & $48.7$ & $46.9$ & $47.8$ & $96.0$& $43.6$ & $\boldsymbol{47.1}$  \\
\hline
\end{tabular}
\caption{Results of the baselines and our $\mathcal{M}^2$ Trans model trained with different joint losses.
For the metrics, BL4, CDr, and BS represent BLEU4, CIDEr-D, and the $\mathrm{F}_1$ score of BERTScore; P, R, $\mathrm{F}_1$ and acc. represent the precision, recall, $\mathrm{F}_1$, and accuracy scores output by the clinical CheXbert labeler, respectively.
For the rewards, $\mathrm{fc}_\mathrm{E}$ and $\mathrm{fc}_\mathrm{EN}$ represent $\mathrm{fact}_\mathrm{ENT}$ and $\mathrm{fact}_\mathrm{ENTNLI}$, respectively.} 
\label{tab:repgen-nlg}
\end{table*}

We used $\mathcal{M}^2$ Trans as our report generation model and used DenseNet-121 \cite{Huang2017} as our image encoder.
We trained $\mathcal{M}^2$ Trans with the following variety of joint losses.
\begin{description}
\item[NLL] $\mathcal{M}^2$ Trans simply optimized with NLL loss as a baseline loss.
\item[NLL+CDr] CIDEr-D and NLL loss is jointly optimized with $\lambda_1 = 0.01$ and $\lambda_2= 0.99$ for the scaling factors.
\item[NLL+BS] The ${\rm F}_1$ score of BERTScore and NLL loss is jointly optimized with $\lambda_1 = 0.01$ and $\lambda_2= 0.99$.
\item[NLL+BS+$\boldsymbol{\mathrm{fc_{E}}}$] $\mathrm{fact_{ENT}}$ is added to NLL+BS with $\lambda_1 = 0.01$, $\lambda_2= 0.495$, and $\lambda_3 = 0.495$.
\item[NLL+BS+$\boldsymbol{\mathrm{fc_{EN}}}$] $\mathrm{fact_{ENTNLI}}$ is added to NLL+BS with $\lambda_1 = 0.01$, $\lambda_2= 0.495$, and $\lambda_3 = 0.495$.
\end{description}

We additionally prepared three previous models that have been tested on MIMIC-CXR.
\begin{description}
\item[TieNet] We reimplemented the model of \citet{Wang2018} consisting of a CNN encoder and an RNN decoder optimized with a multi-task setting of language generation and image classification.
\item[$\boldsymbol{\mathrm{CNN}}$-$\boldsymbol{\mathrm{RNN}^2}$] We reimplemented the model of \citet{Liu2019} consisting of a CNN encoder and a hierarchical RNN decoder optimized with CIDEr and Clinically Coherent Reward which is a reward based on the clinical metrics.
\item[R2Gen] The model of \citet{Chen2020} with a CNN encoder and a memory-driven Transformer optimized with NLL loss. We used the publicly available official code and its checkpoint as its implementation.
\end{description}
For reproducibility, we include model configurations and training details in Appendix B.

\section{Results and Discussions}

\subsection{Evaluation with NLG Metrics and Clinical Metrics}

Table \ref{tab:repgen-nlg} shows the results of the baselines\footnote{These MIMIC-CXR scores have some gaps from the previously reported values with some possible reasons. 
First, TieNet and $\mathrm{CNN}$-$\mathrm{RNN}^2$ in \citet{Liu2019} are evaluated on a pre-release version of MIMIC-CXR. 
Second, we used report-level evaluation for all models, but \citet{Chen2020} tested R2Gen using image-level evaluation.} and $\mathcal{M}^2$ Trans optimized with the five different joint losses.
We find that the best result for a metric or a reward is achieved when that metric or reward is used directly in the optimization objective. 
Notably, for the proposed factual rewards, the increases of $+3.6$ $\rm{fact}_{ENT}$ and $+4.9$ $\rm{fact}_{ENTNLI}$ are observed on MIMIC-CXR with $\mathcal{M}^2$ Trans when compared against $\mathcal{M}^2$ Trans w/ BS.
For the clinical metrics, the best recalls and $\rm{F}_1$ scores are obtained with $\mathcal{M}^2$ Trans using $\rm{fact}_{ENT}$ as a reward, achieving a substantial $+22.1$ increase ($\Delta +63.9\%$) in $\rm{F}_1$ score against the best baseline R2Gen. 
We further find that using $\rm{fact}_{ENTNLI}$ as a reward leads to higher precision and accuracy compared to $\rm{fact}_{ENT}$ with decreases in the recalls.
The best precisions and accuracies were obtained in the baseline $\mathrm{CNN}$-$\mathrm{RNN}^2$.
This is not surprising since this model directly optimizes the clinical  metrics with its Clinically Coherent Reward.
However, this model is strongly optimized against precision resulting in the low recalls and $\mathrm{F}_1$ scores.

The results of $\mathcal{M}^2$ Trans without the proposed rewards and BERTScore reveal the strength of $\mathcal{M}^2$ Trans and the inadequacy of NLL loss and CIDEr for factual completeness and consistency.
$\mathcal{M}^2$ Trans w/ NLL shows strong improvements in the clinical metrics against R2Gen.
These improvements are a little surprising since both models are Transformer-based models and are optimized with NLL loss.
We assume that these improvements are due to architecture differences such as memory matrices in the encoder of $\mathcal{M}^2$ Trans.
The difference between NLL and NLL+CDr on $\mathcal{M}^2$ Trans indicates that NLL and CIDEr are unreliable for factual completeness and consistency.

\begin{figure*}[t]
\centering
\includegraphics[width=2.08\columnwidth]{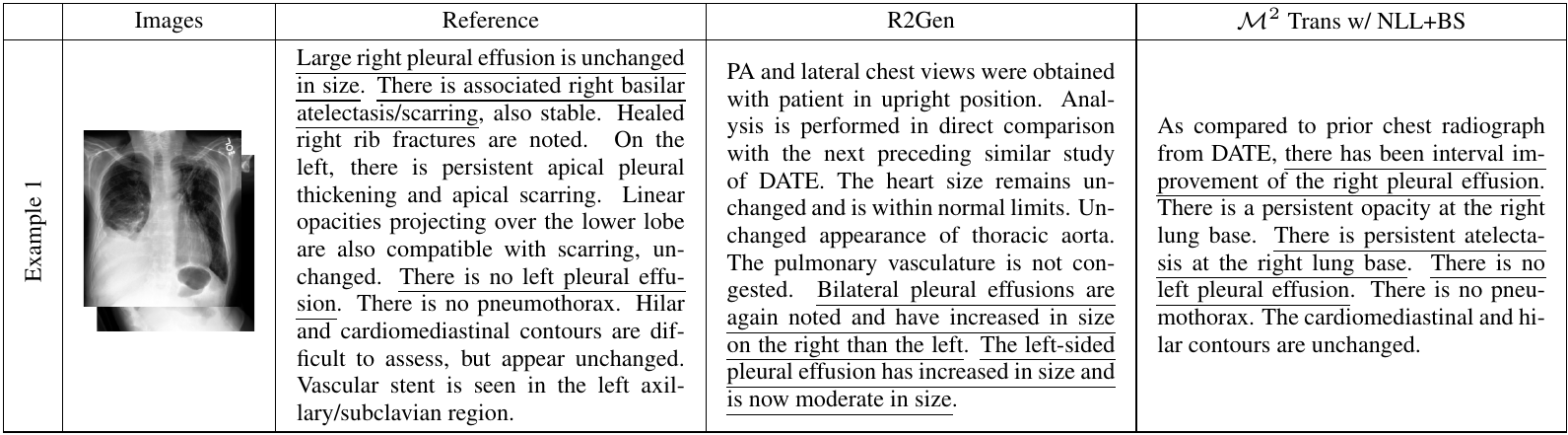} 
\caption{An example of radiology reports generated by R2Gen and by the proposed model with the optimization integrating BERTScore. Repeated sentences are removed from the example to improve readability.}
\label{fig:ex1}
\end{figure*}

\subsection{Human Evaluation}

We performed a human evaluation to further confirm whether the generated radiology reports are factually complete and consistent.
Following prior studies of radiology report summarization \cite{Zhang2020a} and image captioning evaluation  \cite{Vedantam2015cider}, we designed a simple human evaluation task.
Given a reference report (R) and two candidate model generated reports (C1, C2), two board-certified radiologists decided whether C1 or C2 is more factually similar to R.
To consider cases when C1 and C2 are difficult to differentiate, we also prepared ``No difference'' as an answer.
We sampled $100$ reports randomly from the test set of MIMIC-CXR for this evaluation.
Since this evaluation is (financially) expensive and there has been no human evaluation between the baseline models, we selected R2Gen as the best previous model and $\mathcal{M}^2$ Trans w/ BS as the most simple proposed model, in order to be able to weakly infer that all of our proposed models are better than all of the baselines.
Table \ref{tab:human-eval} shows the result of the evaluation.
The majority of the reports were labeled ``No difference'' but the proposed approach received three times as much preference as the baseline.

There are two main reasons why ``No difference'' was frequent in human evaluation. First, we found that a substantial portion of the examples were normal studies (no abnormal observations), which leads to generated reports of similar quality from both models. Second, in some reports with multiple abnormal observations, both models made mistakes on a subset of these observations, making it difficult to decide which model output was better.

\begin{table}[t]
\small
\centering
\begin{tabular}{c | c | c}
\hline
$\mathcal{M}^2$ Trans w/ BS & R2Gen & No \\
\textit{Proposed (simple)} & \cite{Chen2020} & difference \\
\hline
$36.5\%$ & $12.0\%$ & $51.5\%$ \\
\hline
\end{tabular}
\caption{The human evaluation result for randomly sampled $100$ reports from the test set of MIMIC-CXR by two board-certified radiologists.}
\label{tab:human-eval}
\end{table}

\subsection{Estimating Clinical Accuracy with Factual Rewards}

The integrations of $\rm{fact}_{ENT}$ and $\rm{fact}_{ENTNLI}$ showed improvements in the clinical metrics.
We further examined whether these rewards can be used to estimate the performance of the clinical metrics to see whether the proposed rewards can be used in an evaluation where a strong clinical information extraction system like CheXbert is not available.
Table \ref{tab:fact-corr} shows Spearman correlations calculated on the generated reports of NLL+BS.
$\mathrm{fact_{ENTNLI}}$  shows the strongest correlation with the clinical accuracy which aligns with the optimization where the best accuracy is obtained with NLL+ BS+$\mathrm{fact_{ENTNLI}}$.
This correlation value is slightly lower than a Spearman correlation which \citet{Maynez2020} observed with NLI for the factual data ($0.264$).
The result suggests the effectiveness of using the factual rewards to estimate the factual completeness and consistency of radiology reports, although the correlations  are still limited, with some room for improvement.

\begin{table}[t]
\small
\centering
\begin{tabular}{l | c}
\hline
Metric & $\rho$ \\
\hline
BLEU4 & $0.092$ \\
CIDEr-D & $0.034$ \\
BERTScore & $0.155$ \\
$\mathrm{fact_{ENT}}$ & $0.196$ \\
$\mathrm{fact_{ENTNLI}}$ & $\boldsymbol{0.255}$ \\
\hline
\end{tabular}
\caption{The Spearman correlations $\rho$ of NLG metrics and factual metrics against clinical accuracy. The strongest correlation among all metrics is shown is bold.}
\label{tab:fact-corr}
\end{table}

\begin{figure*}[t]
\centering
\includegraphics[width=2.08\columnwidth]{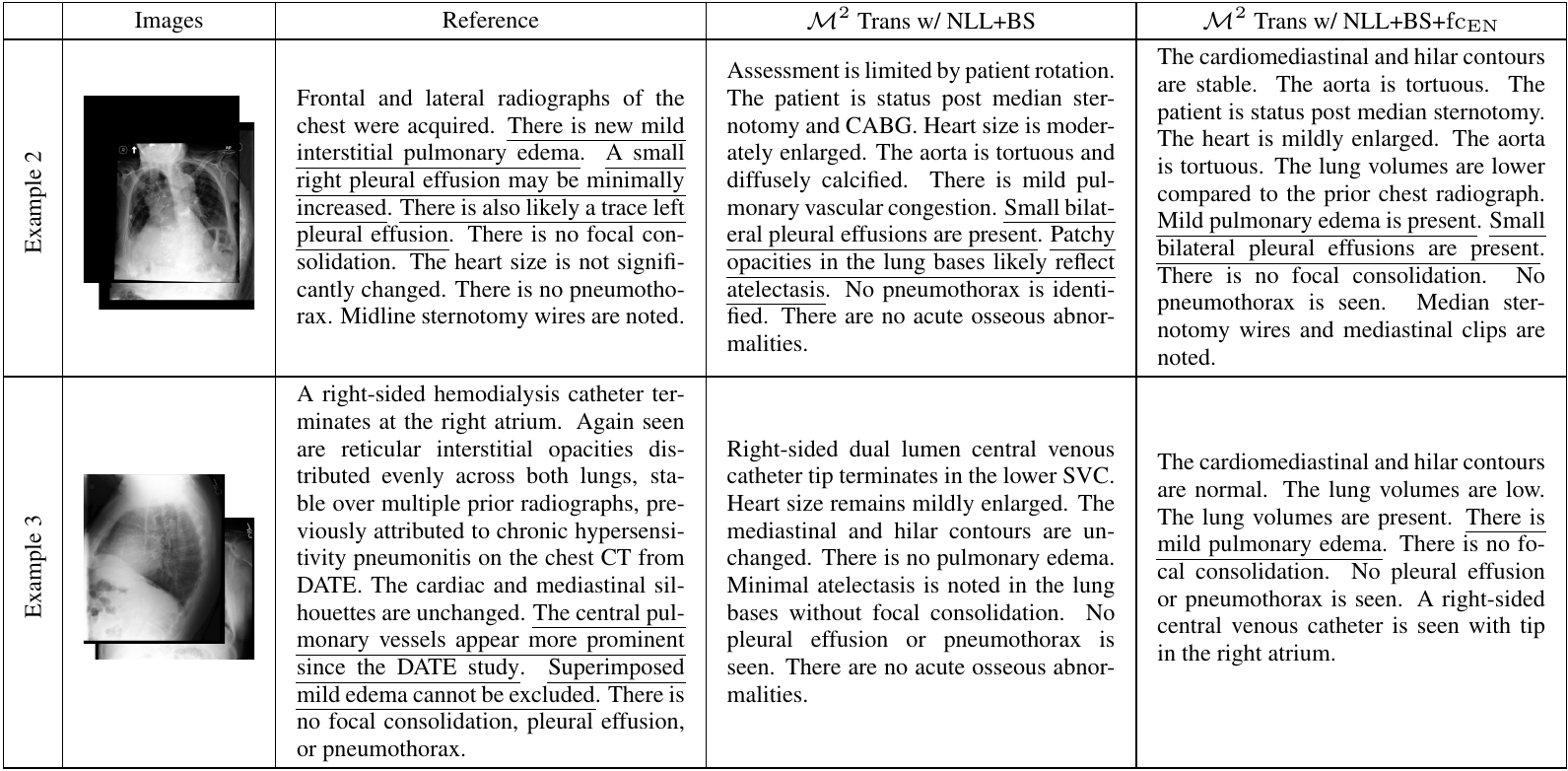} 
\caption{Examples of radiology reports generated by the proposed model with the optimization integrating BERTScore and $\mathrm{fact_{ENTNLI}}$. Repeated sentences are removed from the examples to improve readability.}
\label{fig:ex23}
\end{figure*}

\subsection{Qualitative Analysis of Improved Clinical Completeness and Consistency}

The evaluation with the clinically findings metrics showed  improved generation performance by integrating BERTScore, $\rm{fact}_{ENT}$, and $\rm{fact}_{ENTNLI}$.
As a qualitative analysis, we examined some of the generated reports to see the improvements.
Example 1 in Figure \ref{fig:ex1} shows the improved factual completeness and consistency with BERTScore.
The \textit{atelectasis} is correctly generated and \textit{left plural effusion} is correctly suppressed with NLL+BS.
Example 2 in Figure \ref{fig:ex23} shows the improved factual completeness with $\mathrm{fact_{ENTNLI}}$.
The \textit{edema} is correctly generated and \textit{atelectasis} is correctly suppressed with NLL+BS+$\mathrm{fc_{EN}}$.
These examples reveal the strength of integrating the three metrics to generate factually complete and consistent reports.

Despite observing large improvements with our model in the clinical finding metrics evaluation, the model is still not complete and some typical factual errors can be found in their generated reports.
For example, Example 3 in Figure \ref{fig:ex23} includes a comparison of an observation against a previous study as ``\ldots appear more prominent since \ldots'' in the reference but our model (or any previous models) can not capture this kind of comparison since the model is not designed to take account the past reports of a patient as input.
Additionally, in this example, \textit{edema} is mentioned with uncertainty as ``cannot be excluded'' in the reference but the generated report with $\rm{fact}_{ENTNLI}$ simply indicates it as ``There is mild pulmonary edema''.

\section{Conclusion}
We proposed two new simple rewards and combined them with a semantic equivalence metric to improve image-to-text radiology report generation systems.
The two new rewards make use of radiology domain entities extracted with a named entity recognizer and a weakly-supervised NLI to capture the factual completeness and consistency of the generated reports.
We further presented a Transformer-based report generation system that directly optimizes these rewards with self-critical reinforcement learning.
On two open datasets, we showed that our system generates reports that are more factually complete and consistent than the baselines and leads to reports with substantially higher scores in clinical  metrics.
The integration of entities and NLI to improve the factual completeness and consistency of generation is not restricted to the domain of radiology reports, and we predict that a similar approach might similarly improve other data-to-text tasks.

\section*{Acknowledgements}
We would like to thank the anonymous reviewers and the members of the Stanford NLP Group for their very helpful comments that substantially improved this paper. 

\bibliography{references}
\bibliographystyle{acl_natbib}

\clearpage

\appendix

\section{Detail of Radiology NLI}
\label{sec:app-nli}

\subsection{Rules \& Examples of Weakly-Supervised Radiology NLI}

We prepared the $6$ rules (E1, N1--N4, and C1) to train the weakly-supervised radiology NLI.
The rules are applied against sentence pairs consisting from premises ($s_1$) and hypotheses ($s_2$) to extract pairs that are in \textit{entailment}, \textit{neutral}, or \textit{contradiction} relation.

\subsubsection*{Entailment Rule: E1}
\begin{enumerate}
\item $s_1$ and $s_2$ are semantically similar.
\item The named entities (NE) of $s_2$ is a subset or equal to the named entities of $s_1$ as $\rm{NE}(\rm{s_2}) \subseteq \rm{NE}(\rm{s_1})$.
\end{enumerate}
We used BERTScore \cite{Zhang2020bertscore} as a similarity metric and set the threshold to $sim(\rm{s_1}, \rm{s_2}) \ge 0.7$\footnote{\textit{distilbert-base-uncased} with the baseline score  is used as the model of BERTScore for a fast comparison and a smooth score scale. We swept the threshold value from $\{ 0.6, 0.7, 0.8, 0.9 \}$ and set it to $0.7$ as a relaxed boundary to balance between accuracy and diversity. }.
The clinical model of Stanza \cite{Zhang2020c} is used to extract \textit{anatomy} entities and \textit{observation} entities.
$s_1$ and $s_2$ are conditioned to be both negated or both non-negated.
The negation is determined with a negation identifier or the existence of \textit{uncertain} entity, using NegBio \cite{Peng2018} as the negation identifier and the clinical model of Stanza is used to extract \textit{uncertain} entities.
$s_2$ is further restricted to include at least $2$ entities as $| \rm{NE}( \rm{s_2} ) | \ge 2$.
These similarity metric, named entity recognition model, and entity number restriction are used in the latter neutral and contradiction rules.
The negation restriction is used in the neutral rules but is not used in the contradiction rule.
The following is an example of a sentence pair that matches E1 with entities in bold:
\begin{description}
\item[$\boldsymbol{s_1}$] The \textbf{heart} is mildly \textbf{enlarged}. 
\item[$\boldsymbol{s_2}$] The \textbf{heart} appears again mild-to-moderately \textbf{enlarged}.
\end{description}

\subsubsection*{Neutral Rule 1: N1}
\begin{enumerate}
\item $s_1$ and $s_2$ are semantically similar.
\item The named entities of $s_1$ is a subset of the named entities of $s_2$ as $\rm{NE}(\rm{s_1}) \subsetneq \rm{NE}(\rm{s_2})$.
\end{enumerate}
Since $s_1$ is a premise, this condition denotes that the counterpart hypothesis has entities that are not included in the premise.
The following is an example of a sentence pair that matches N1 with entities in bold:
\begin{description}
\item[$\boldsymbol{s_1}$] There is no \textbf{pulmonary edema} or definite \textbf{consolidation}.
\item[$\boldsymbol{s_2}$] There is no focal \textbf{consolidation}, \textbf{pleural effusion}, or \textbf{pulmonary edema}.
\end{description}

\subsubsection*{Neutral Rule 2: N2}
\begin{enumerate}
\item The named entities of $s_1$ are equal to the named entities of $s_2$ as $\rm{NE}(\rm{s_1})=\rm{NE}(\rm{s_2})$.
\item The anatomy modifiers ($\rm{NE}_{mod}$) of $s_1$ include an antonym (ANT) of the anatomy modifier of $s_2$ as $\rm{NE}_{\rm mod} (s_1) \cap ANT( \rm{NE}_{\rm mod} (s_2) ) \neq \emptyset$.
\end{enumerate}
Anatomy modifiers are extracted with the clinical model of Stanza and antonyms are decided using WordNet \cite{Fellbaum1998}.
Antonyms in anatomy modifiers are considered in this rule to differentiate experessions like \textit{left vs right} and \textit{upper vs lower}.
The following is an example of a sentence pair that matches N2 with antonyms in bold:
\begin{description}
\item[$\boldsymbol{s_1}$] Moreover, a small \textbf{left} pleural effusion has newly occurred.
\item[$\boldsymbol{s_2}$] Small \textbf{right} pleural effusion has worsened.
\end{description}

\subsubsection*{Neutral Rule 3: N3}
\begin{enumerate}
\item The named entity types ($\rm{NE}_{type}$) of $s_1$ are equal to the named entity types of $s_2$ as $\rm{NE}_{\rm type}(\rm{s_1}) = \rm{NE}_{\rm type}(\rm{s_2})$.
\item The named entities of $s_1$ is different from the named entities of $s_2$ as $\rm{NE}(\rm{s_1}) \cap \rm{NE}(\rm{s_2}) = \emptyset$.
\end{enumerate}
Specific entity types that we used are \textit{anatomy} and \textit{observation}.
This rule ensures that $s_1$ and $s_2$ have related but different entities in same types.
The following is an example of a sentence pair that matches N3 with entities in bold:
\begin{description}
\item[$\boldsymbol{s_1}$] There is minimal bilateral \textbf{lower lobe atelectasis}.
\item[$\boldsymbol{s_2}$] The \textbf{cardiac silhouette} is moderately \textbf{enlarged}.
\end{description}

\subsubsection*{Neutral Rule 4: N4}
\begin{enumerate}
\item The named entities of $s_1$ are equal to the named entities of $s_2$ as $\rm{NE}(\rm{s_1})=\rm{NE}(\rm{s_2})$.
\item $s_1$ and $s_2$ include observation keywords (KEY) that belong to different groups as $\rm{KEY}(s_1) \neq \rm{KEY}(s_2)$.
\end{enumerate}
The groups of observation keywords are setup following the observation keywords of CheXpert labeler \cite{Irvin2019}.
Specifically, ${\rm G1} = \{ normal, unremarkable \}$, ${\rm G2} = \{ stable, unchanged \}$, and ${\rm G3} = \{ clear \}$ are used to determine words included in different groups as \textit{neutral} relation.
The following is an example of a sentence pair that matches N4 with keywords in bold:
\begin{description}
\item[$\boldsymbol{s_1}$] \textbf{Normal} cardiomediastinal silhouette.
\item[$\boldsymbol{s_2}$] Cardiomediastinal silhouette is \textbf{unchanged}.
\end{description}

\subsubsection*{Contradiction Rule: C1}
\begin{enumerate}
\item The named entities of $s_1$ is a subset or equal to the named entities of $s_2$ as $\rm{NE}(\rm{s_2}) \subseteq \rm{NE}(\rm{s_1})$.
\item $s_1$ or $s_2$ is a negated sentence.
\end{enumerate}
Negation is determined with the same approach as E1.
The following is an example of a sentence pair that matches C1 with entities in bold:
\begin{description}
\item[$\boldsymbol{s_1}$] There are also small bilateral \textbf{pleural effusions}.
\item[$\boldsymbol{s_2}$] No \textbf{pleural effusions}.
\end{description}

\subsection{Validation and Test Datasets of Radiology NLI}

We sampled $480$ sentence pairs that satisfy the following conditions from the validation section of MIMIC-CXR:
\begin{enumerate}
\item Two sentences ($s_1$ and $s_2$) have ${\rm BERTScore (s_1, s_2)} \ge 0.5$.
\item MedNLI labels are equally distributed over three labels: \textit{entailment}, \textit{neutral}, and \textit{contradiction}\footnote{We used the baseline BERT model of \citeauthor{Wu2019} \shortcite{Wu2019} to assign MedNLI labels to the pairs.}.
\end{enumerate}
These conditions are introduced to reduce \textit{neutral} pairs since most pairs will be \textit{neutral} with random sampling.
The sampled pairs are annotated twice swapping its premise and hypothesis by two experts: one medical expert and one NLP expert.
For pairs that the two annotators disagreed, its labels are decided by a discussion with one additional NLP expert.
The resulting $960$ bidirectional pairs are splitted in half resulting in $480$ pairs for a validation set and $480$ pairs for a test set.

\subsection{Configuration of Radiology NLI Model}

We used \textit{bert-base-uncased} as a pre-trained BERT model and further fine-tuned it on MIMIC-III \cite{Johnson2016} radiology reports with a masked language modeling loss for $8$ epochs. 
The model is further optimized on the training data with a classification negative log likelihood loss.
We used Adam \cite{Kingma2015} as an optimization method with $\beta_1=0.9$, $\beta_2=0.999$, batch size of $16$, and the gradient clipping norm of $5.0$.
The learning rate is set to $lr = 1e^{-5}$ by running a preliminary experiment with $lr = \{ 1e^{-5}, 2e^{-5} \}$.
The model is optimized for the maximum of $20$ epochs and a validation accuracy is used to decide a model checkpoint that is used to evaluate the test set.
We trained the model with a single Nvidia Titan XP taking approximately $2$ hours to complete $20$ epochs.

\section{Configurations of Radiology Report Generation Models}
\label{sec:app-gen-config}

\subsection{$\boldsymbol{\mathcal{M}^2}$ Trans}
We used DenseNet-121 \cite{Huang2017} as a CNN image feature extractor and pre-trained it on CheXpert dataset with the $14$-class classification setting.
We used GloVe \cite{Pennington2014} to pre-train text embeddings and the pre-trainings were done on a training set with the embedding size of $512$.
The parameters of the model is set up to the dimensionality of $512$, the number of heads to $8$, and the number of memory vector to $40$.
We set the number of Transformer layer to $n_{layer}=1$ by running a preliminary experiment with $n_{layer} = \{1, 2, 3\}$.
The model is first trained against NLL loss using the learning rate scheduler of Transformer \cite{Devlin2019} with the warm-up steps of $20000$ and is further optimized with a  joint loss with the fixed learning rate of $5e^{-6}$.
Adam is used as an optimization method with $\beta_1=0.9$ and $\beta_2=0.999$.
The batch size is set to $48$ for NLL loss and $24$ for the joint losses.
For $\lambda_*$, we first swept the optimal value of $\lambda_1$ from $\{ 0.03, 0.02, 0.01, 0.001 \}$ using the development set.
We have restricted $\lambda_2$ and $\lambda_3$ to have equal values in our experiments and constrined that all $\lambda_*$ values sum up to $1.0$.
The model is trained with NLL loss for $32$ epochs and further trained for $32$ epochs with a joint loss.
Beam search with the beam size of $4$ is used to decode texts when evaluating the model against a validation set or a test set.
We trained the model with a single Nvidia Titan XP taking approximately $10$ days to complete its optimization.

\begin{table*}[t]
\small
\centering
\def\arraystretch{1.05} 
\newcommand\T{\rule{0pt}{5ex}} 
\newcommand\B{\rule[-3.5ex]{0pt}{0pt}} 
\begin{tabular}{ll | cccc | cccc}
\hline
& & \multicolumn{4}{c|}{MIMIC-CXR} & \multicolumn{4}{c}{Open-i} \\
\cline{3-10}
\multicolumn{2}{l|}{\makecell[l]{Observation  \\ (\# MIMIC-CXR / \# Open-i)}}
& \makecell[c]{R2Gen}
& \makecell[c]{NLL + \\ BS}
& \makecell[c]{NLL + \\ BS + \\$\mathrm{fc_{E}}$}
& \makecell[c]{NLL + \\ BS + \\$\mathrm{fc_{EN}}$}
& \makecell[c]{R2Gen}
& \makecell[c]{NLL + \\ BS}
& \makecell[c]{NLL + \\ BS + \\$\mathrm{fc_{E}}$}
& \makecell[c]{NLL + \\ BS + \\$\mathrm{fc_{EN}}$} \T\B\\
\hhline{==|====|====}
& P & $41.2$ & $46.3$ & $46.3$ & $50.3$ & $27.0$ & $47.7$ & $44.0$ & $48.7$ \\
Micro Average & R & $29.8$ & $67.5$ & $73.2$ & $65.1$ & $17.3$ & $46.6$ & $53.5$ & $46.9$ \\
($2713$ / $654$) & $\mathrm{F}_1$ & $34.6$ & $54.9$ & $56.7$ & $56.7$ & $21.1$ & $47.2$ & $48.3$ & $47.8$ \\
& acc. & $73.9$ & $74.4$ & $74.2$ & $77.1$ & $94.9$ & $95.9$ & $95.5$ & $96.0$ \\
\hhline{==|====|====}
& P & $35.4$ & $39.9$ & $37.9$ & $40.6$ & $35.6$ & $44.0$ & $35.8$ & $39.4$ \\
Atelectasis & R & $27.8$ & $67.4$ & $80.5$ & $76.2$ & $7.4$ & $35.6$ & $47.7$ & $45.4$ \\
($604$ / $216$) & $\mathrm{F}_1$ & $31.1$ & $50.2$ & $51.6$ & $53.0$ & $12.3$ & $39.4$ & $40.9$ & $42.2$ \\
& acc. & $68.3$ & $65.5$ & $61.1$ & $65.2$ & $93.1$ & $92.9$ & $91.1$ & $91.9$ \\
\hline
& P & $32.4$ & $35.8$ & $34.3$ & $37.5$ & $24.5$ & $56.6$ & $57.3$ & $60.0$ \\
Cardiomegaly & R & $53.5$ & $73.3$ & $81.3$ & $61.3$ & $33.8$ & $55.6$ & $55.6$ & $46.7$ \\
($535$ / $225$) & $\mathrm{F}_1$ & $40.4$ & $48.1$ & $48.2$ & $46.6$ & $28.4$ & $56.1$ & $56.4$ & $52.5$ \\
& acc. & $64.0$ & $63.9$ & $60.2$ & $67.9$ & $88.5$ & $94.1$ & $94.2$ & $94.3$ \\
\hline
& P & $14.3$ & $10.5$ & $19.6$ & $19.2$ & $0.0$ & $10.9$ & $15.2$ & $14.3$ \\
Consolidation & R  & $7.0$ & $18.5$ & $5.7$ & $3.2$ & $0.0$ & $26.3$ & $26.3$ & $5.3$ \\
($157$ / $19$) & $\mathrm{F}_1$ & $9.4$ & $13.4$ & $8.9$ & $5.5$ & $0.0$ & $15.4$ & $19.2$ & $7.7$ \\
& acc. & $91.0$ & $84.0$ & $92.1$ & $92.6$ & $99.0$ & $98.4$ & $98.7$ & $99.3$ \\
\hline
& P & $55.3$ & $59.7$ & $56.0$ & $65.6$ & $10.0$ & $39.0$ & $30.9$ & $41.4$ \\
Edema  & R & $24.3$ & $59.2$ & $69.9$ & $52.7$ & $4.0$ & $30.7$ & $50.7$ & $32.0$ \\
($645$ / $75$) & $\mathrm{F}_1$ & $33.8$ & $59.5$ & $62.2$ & $58.5$ & $5.7$ & $34.3$ & $38.4$ & $36.1$ \\
& acc. & $73.8$ & $77.8$ & $76.6$ & $79.4$ & $97.0$ & $97.4$ & $96.3$ & $97.5$ \\
\hline
& P & $76.2$ & $67.2$ & $68.2$ & $65.9$ & $85.7$ & $54.3$ & $59.4$ & $56.0$ \\
Pleural Effusion& R  & $24.1$ & $80.6$ & $78.5$ & $82.0$ & $15.1$ & $63.0$ & $66.4$ & $66.4$ \\
($772$ / $119$) & $\mathrm{F}_1$ & $36.6$ & $73.3$ & $73.0$ & $73.1$ & $25.7$ & $58.4$ & $62.7$ & $60.8$ \\
& acc. & $72.6$ & $80.7$ & $80.9$ & $80.1$ & $96.9$ & $96.8$ & $97.2$ & $96.9$ \\
\hline
\end{tabular}
\caption{The detailed results of R2Gen, $\mathcal{M}^2$ Trans w/ BS, $\mathcal{M}^2$ Trans w/ BS+$\mathrm{{fc}_{E}}$, and $\mathcal{M}^2$ Trans w/ BS+$\mathrm{{fc}_{EN}}$ for the $5$ observations. P is precision, R is recall, and acc. is accuracy. \#MIMIC- CXR and \#Open-i are the numbers of times that a corresponding observation has appeared as positive in the test set of MIMIC-CXR and Open-i, respectively. }
\label{tab:repgen-detailed}
\end{table*}

\begin{table*}[t]
\small
\centering
\def\arraystretch{1.05} 
\newcommand\T{\rule{0pt}{5ex}} 
\newcommand\B{\rule[-3.5ex]{0pt}{0pt}} 
\begin{tabular}{ll | cccc | cccc}
\hline
& & \multicolumn{4}{c|}{MIMIC-CXR} & \multicolumn{4}{c}{Open-i} \\
\cline{3-10}
\multicolumn{2}{l|}{\makecell[l]{Observation  \\ (\# MIMIC-CXR / \# Open-i)}}
& \makecell[c]{R2Gen}
& \makecell[c]{NLL + \\ BS}
& \makecell[c]{NLL + \\ BS + \\$\mathrm{fc_{E}}$}
& \makecell[c]{NLL + \\ BS + \\$\mathrm{fc_{EN}}$}
& \makecell[c]{R2Gen}
& \makecell[c]{NLL + \\ BS}
& \makecell[c]{NLL + \\ BS + \\$\mathrm{fc_{E}}$}
& \makecell[c]{NLL + \\ BS + \\$\mathrm{fc_{EN}}$} \T\B\\
\hhline{==|====|====}
& P & $37.6$ & $46.0$ & $46.0$ & $49.9$ & $16.7$ & $46.3$ & $42.7$ & $47.8$ \\
Micro Average & R & $29.1$ & $67.2$ & $72.9$ & $64.6$ & $17.1$ & $45.8$ & $52.5$ & $46.3$ \\
($2713$ / $654$) & $\mathrm{F}_1$ & $32.8$ & $54.6$ & $56.4$ & $56.3$ & $16.9$ & $46.1$ & $47.1$ & $47.0$ \\
& acc. & $74.6$ & $74.2$ & $74.0$ & $76.8$ & $93.4$ & $95.8$ & $95.4$ & $95.9$\\
\hhline{==|====|====}
& P & $35.6$ & $39.9$ & $37.8$ & $40.6$ & $33.3$ & $42.9$ & $34.7$ & $38.2$ \\
Atelectasis & R & $23.9$ & $67.6$ & $80.6$ & $76.4$ & $7.1$ & $35.4$ & $47.2$ & $44.8$ \\
($604$ / $216$) & $\mathrm{F}_1$ & $28.6$ & $50.2$ & $51.5$ & $53.0$ & $11.7$ & $38.8$ & $40.0$ & $41.2$ \\
& acc. & $72.1$ & $65.6$ & $61.1$ & $65.3$ & $93.2$ & $92.9$ & $91.0$ & $91.9$ \\
\hline
& P & $28.6$ & $36.1$ & $34.6$ & $37.6$ & $20.4$ & $55.2$ & $55.5$ & $60.0$ \\
Cardiomegaly & R & $49.2$ & $72.8$ & $81.1$ & $60.5$ & $31.3$ & $53.5$ & $53.0$ & $45.7$ \\
($535$ / $225$) & $\mathrm{F}_1$ & $36.2$ & $48.2$ & $48.5$ & $46.4$ & $24.7$ & $54.3$ & $54.2$ & $51.9$ \\
& acc. & $63.0$ & $63.8$ & $60.0$ & $67.6$ & $86.8$ & $93.8$ & $93.8$ & $94.2$ \\
\hline
& P & $10.7$ & $10.5$ & $19.6$ & $19.2$ & $1.1$ & $10.9$ & $14.7$ & $14.3$ \\
Consolidation & R  & $9.4$ & $17.8$ & $5.5$ & $3.1$ & $10.5$ & $26.3$ & $26.3$ & $5.3$ \\
($157$ / $19$) & $\mathrm{F}_1$ & $10.0$ & $13.2$ & $8.6$ & $5.3$ & $2.0$ & $15.4$ & $18.9$ & $7.7$ \\
& acc. & $88.4$ & $83.7$ & $91.9$ & $92.4$ & $94.0$ & $98.4$ & $98.7$ & $99.3$ \\
\hline
& P & $49.0$ & $58.7$ & $54.9$ & $64.3$ & $5.8$ & $35.0$ & $30.1$ & $39.7$ \\
Edema  & R & $29.1$ & $59.0$ & $69.5$ & $52.3$ & $4.1$ & $28.8$ & $50.7$ & $31.5$ \\
($645$ / $75$) & $\mathrm{F}_1$ & $36.5$ & $58.8$ & $61.4$ & $57.7$ & $4.8$ & $31.6$ & $37.8$ & $35.1$ \\
& acc. & $74.5$ & $77.6$ & $76.2$ & $79.2$ & $96.4$ & $97.3$ & $96.3$ & $97.5$ \\
\hline
& P & $75.6$ & $66.5$ & $67.6$ & $65.2$ & $63.3$ & $53.2$ & $57.5$ & $54.6$ \\
Pleural Effusion& R  & $23.4$ & $80.3$ & $78.2$ & $81.6$ & $16.4$ & $63.8$ & $66.4$ & $66.4$ \\
($772$ / $119$) & $\mathrm{F}_1$ & $35.8$ & $72.7$ & $72.5$ & $72.5$ & $26.0$ & $58.0$ & $61.6$ & $59.9$ \\
& acc. & $75.1$ & $80.3$ & $80.6$ & $79.8$ & $96.8$ & $96.8$ & $97.1$ & $96.9$ \\
\hline
\end{tabular}
\caption{The detailed results of R2Gen, $\mathcal{M}^2$ Trans w/ BS, $\mathcal{M}^2$ Trans w/ BS+$\mathrm{{fc}_{E}}$, and $\mathcal{M}^2$ Trans w/ BS+$\mathrm{{fc}_{EN}}$ for the $5$ observations evaluated with CheXpert instead of CheXbert.}
\label{tab:repgen-detailed2}
\end{table*}

\begin{table*}[t]
\small
\centering
\def\arraystretch{1.05} 
\newcommand\T{\rule{0pt}{5ex}} 
\newcommand\B{\rule[-3.5ex]{0pt}{0pt}} 
\begin{tabular}{ll | cccc | cccc}
\hline
& & \multicolumn{4}{c|}{MIMIC-CXR} & \multicolumn{4}{c}{Open-i} \\
\cline{3-10}
\multicolumn{2}{l|}{\makecell[l]{Observation  \\ (\# MIMIC-CXR / \# Open-i)}}
& \makecell[c]{R2Gen}
& \makecell[c]{NLL + \\ BS}
& \makecell[c]{NLL + \\ BS + \\$\mathrm{fc_{E}}$}
& \makecell[c]{NLL + \\ BS + \\$\mathrm{fc_{EN}}$}
& \makecell[c]{R2Gen}
& \makecell[c]{NLL + \\ BS}
& \makecell[c]{NLL + \\ BS + \\$\mathrm{fc_{E}}$}
& \makecell[c]{NLL + \\ BS + \\$\mathrm{fc_{EN}}$} \T\B\\
\hhline{==|====|====}
& P & $4.4$ & $5.1$ & $4.8$ & $4.6$ & $3.8$ & $0.7$ & $2.0$ & $4.0$ \\
Enlarged Cardiomediastinum & R & $19.8$ & $50.5$ & $19.8$ & $47.7$ & $16.7$ & $4.2$ & $4.2$ & $20.8$ \\
($111$ / $24$) & $\mathrm{F}_1$ & $7.1$ & $9.3$ & $7.7$ & $8.4$ & $6.3$ & $1.2$ & $2.7$ & $6.7$ \\
& acc. & $75.6$ & $53.4$ & $77.5$ & $50.6$ & $96.4$ & $95.1$ & $97.8$ & $95.8$ \\
\hline
& P & $0.0$ & $40.0$ & $10.7$ & $26.1$ & $0.0$ & $0.0$ & $0.0$ & $3.1$ \\
Fracture & R & $0.0$ & $3.6$ & $5.4$ & $10.7$ & $0.0$ & $0.0$ & $0.0$ & $2.3$ \\
($56$ / $43$) & $\mathrm{F}_1$ & $0.0$ & $6.6$ & $7.1$ & $15.2$ & $0.0$ & $0.0$ & $0.0$ & $2.7$ \\
& acc. & $97.6$ & $97.6$ & $96.7$ & $97.1$ & $98.7$ & $98.6$ & $98.1$ & $97.8$ \\
\hline
& P & $37.5$ & $33.3$ & $22.2$ & $44.4$ & $25.0$ & $0.0$ & $0.0$ & $66.7$ \\
Lung Lesion & R  & $3.1$ & $1.0$ & $2.1$ & $4.1$ & $1.1$ & $0.0$ & $0.0$ & $4.5$ \\
($97$ / $89$) & $\mathrm{F}_1$ & $5.7$ & $2.0$ & $3.8$ & $7.5$ & $2.2$ & $0.0$ & $0.0$ & $8.4$ \\
& acc. & $95.8$ & $95.8$ & $95.7$ & $95.8$ & $97.3$ & $97.3$ & $97.3$ & $97.4$ \\
\hline
& P & $44.5$ & $48.8$ & $53.5$ & $54.9$ & $43.1$ & $50.5$ & $57.8$ & $41.1$ \\
Lung Opacity & R & $29.9$ & $41.6$ & $10.4$ & $26.6$ & $8.1$ & $29.1$ & $7.6$ & $22.1$ \\
($798$ / $344$) & $\mathrm{F}_1$ & $35.8$ & $44.9$ & $17.4$ & $35.8$ & $13.7$ & $36.9$ & $13.4$ & $28.7$ \\
& acc. & $63.5$ & $65.3$ & $66.5$ & $67.6$ & $89.4$ & $89.7$ & $89.9$ & $88.7$ \\
\hline
& P & $31.4$ & $44.4$ & $49.8$ & $48.8$ & $78.1$ & $80.8$ & $82.1$ & $81.7$ \\
No Finding & R  & $43.9$ & $35.9$ & $41.7$ & $39.9$ & $84.1$ & $93.4$ & $91.5$ & $88.4$ \\
($396$ / $2319$) & $\mathrm{F}_1$ & $36.6$ & $39.7$ & $45.4$ & $43.9$ & $81.0$ & $86.6$ & $86.5$ & $84.9$ \\
& acc. & $74.4$ & $81.6$ & $83.1$ & $82.8$ & $72.6$ & $80.0$ & $80.2$ & $78.2$ \\
\hline
& P & $0.0$ & $0.0$ & $0.0$ & $0.0$ & $0.0$ & $0.0$ & $0.0$ & $0.0$ \\
Pleural Other & R  & $0.0$ & $0.0$ & $0.0$ & $0.0$ & $0.0$ & $0.0$ & $0.0$ & $0.0$ \\
($39$ / $29$) & $\mathrm{F}_1$ & $0.0$ & $0.0$ & $0.0$ & $0.0$ & $0.0$ & $0.0$ & $0.0$ & $0.0$ \\
& acc. & $98.1$ & $98.3$ & $98.3$ & $98.3$ & $99.1$ & $99.1$ & $99.1$ & $99.1$ \\
\hline
& P & $42.1$ & $62.7$ & $62.1$ & $0.0$ & $27.6$ & $31.3$ & $38.6$ & $0.0$ \\
Pneumonia & R  & $15.8$ & $16.3$ & $17.0$ & $0.0$ & $7.3$ & $19.3$ & $24.8$ & $0.0$ \\
($424$ / $109$) & $\mathrm{F}_1$ & $23.0$ & $25.8$ & $26.7$ & $0.0$ & $11.6$ & $23.9$ & $30.2$ & $0.0$ \\
& acc. & $80.9$ & $83.1$ & $83.1$ & $81.9$ & $96.3$ & $96.0$ & $96.3$ & $96.7$ \\
\hline
& P & $60.0$ & $28.7$ & $37.0$ & $50.0$ & $100.0$ & $40.0$ & $0.0$ & $100.0$ \\
Pneumothorax & R  & $3.8$ & $34.6$ & $12.8$ & $10.3$ & $6.7$ & $13.3$ & $0.0$ & $13.3$ \\
($78$ / $15$) & $\mathrm{F}_1$ & $7.2$ & $31.4$ & $19.0$ & $17.0$ & $12.5$ & $20.0$ & $0.0$ & $23.5$ \\
& acc. & $96.7$ & $95.0$ & $96.4$ & $96.7$ & $99.6$ & $99.5$ & $99.6$ & $99.6$ \\
\hline
& P & $52.2$ & $50.8$ & $53.2$ & $49.0$ & $10.0$ & $16.2$ & $19.7$ & $13.1$ \\
Support Devices & R  & $68.9$ & $83.5$ & $78.7$ & $89.7$ & $12.2$ & $43.9$ & $36.6$ & $56.1$ \\
($624$ / $41$) & $\mathrm{F}_1$ & $59.4$ & $63.2$ & $63.5$ & $63.3$ & $11.0$ & $23.7$ & $25.6$ & $21.3$ \\
& acc. & $75.0$ & $74.1$ & $75.9$ & $72.4$ & $97.6$ & $96.5$ & $97.4$ & $94.9$ \\
\hline
\end{tabular}
\caption{The detailed results of R2Gen, $\mathcal{M}^2$ Trans w/ BS, $\mathcal{M}^2$ Trans w/ BS+$\mathrm{{fc}_{E}}$, and $\mathcal{M}^2$ Trans w/ BS+$\mathrm{{fc}_{EN}}$ for the remaining $9$ observations. }
\label{tab:repgen-detailed3}
\end{table*}

\subsection{TieNet}
We used ResNet-50 as a CNN image feature extractor with default ImageNet pre-trained weights.
We used GloVe to pre-train text embeddings with the same configuration as $\mathcal{M}^2$ Trans.
The parameters of the model is set up to the LSTM dimension of $256$ and the number of global attentions to $5$.
The combination of NLL loss and the multi-label classification loss is used as its joint loss with the balance parameter $\alpha=0.85$.
The model is trained against the joint loss using a linear rate scheduler with the initial learning rate of $1e^{-4}$ and the multiplication of $0.5$ per $8$ epochs.
The batch size is set to $32$ and the model is trained with the joint loss for $32$ epochs.
Adam is used as an optimization method with $\beta_1=0.9$ and $\beta_2=0.999$.
Beam search with the beam size of $4$ is used to decode texts.
We trained the model with a single Nvidia Titan XP taking approximately $2$ days to complete its optimization.

\subsection{$\boldsymbol{\mathrm{CNN}}$-$\boldsymbol{\mathrm{RNN^2}}$}
We used DenseNet-121 as a CNN image feature extractor with default ImageNet pre-trained weights.
We used GloVe to pre-train text embeddings with the same configuration as $\mathcal{M}^2$ Trans.
The parameters of the model is set up to the LSTM dimension of $256$.
We modified an information extraction system from CheXpert to CheXbert to improve the training speed of this model.
The combination of CIDEr and Clinically Coherent Reward is used as its joint loss with the balance parameter $\lambda=10.0$.
The model is first trained against NLL loss using a linear rate scheduler with the initial learning rate of $1e^{-4}$ and the multiplication of $0.5$ per $8$ epochs.
The model is further optimized with the joint loss with the fixed learning rate of $5e^{-6}$.
Adam is used as an optimization method with $\beta_1=0.9$ and $\beta_2=0.999$.
The batch size is set to $32$ for the NLL loss and $24$ for the joint losses.
The model is trained with NLL loss for $32$ epochs and further trained for $32$ epochs with the joint loss.
Beam search with the beam size of $4$ is used to decode texts.
We trained the model with a single Nvidia Titan XP taking approximately $11$ days to complete its optimization.

\section{Detailed Result of Clinical Metrics}
\label{sec:app-cfm}

Table \ref{tab:repgen-detailed} shows the detailed results of the clinical metrics for R2Gen, $\mathcal{M}^2$ Trans w/ BS, $\mathcal{M}^2$ Trans w/ BS+$\mathrm{{fc}_{E}}$, and $\mathcal{M}^2$ Trans w/ BS+$\mathrm{{fc}_{EN}}$.
In most cases, the best $\mathrm{F}_1$ scores are observed in the cases when $\mathrm{fact_{ENT}}$ or $\mathrm{fact_{ENTNLI}}$ is included in the joint losses.
\textit{Consolidation} is one exception where the best precisions, recalls, and $\mathrm{F}_1$ scores vary among the joint losses.
We assume this is due to the infrequent appearance of \textit{consolidation} in both MIMIC-CXR and Open-i.
For comparison against some past studies, we show the detailed results when CheXpert is used instead of CheXbert in Table \ref{tab:repgen-detailed2}.
Since CheXbert is more or equally accurate for most observations than CheXpert, the scores in Table \ref{tab:repgen-detailed2} follow similar trends against ones in Table \ref{tab:repgen-detailed}.
Table \ref{tab:repgen-detailed3} shows the detailed results for the $9$ remaining observations that are defined in CheXpert.
Note that many of these observations are infrequent and have relatively weaker and unstable extraction performances compared to the $5$ observations in Table \ref{tab:repgen-detailed}.

\end{document}